\documentclass[final]{cvpr}

\usepackage{times}
\usepackage{epsfig}
\usepackage{graphicx}
\usepackage{amsmath}
\usepackage{amssymb}

\usepackage[pagebackref=true,breaklinks=true,colorlinks,bookmarks=false]{hyperref}

\usepackage{mathtools}
\usepackage{multirow}
\usepackage{amsfonts}
\usepackage{bbm}

\newcommand{\SOTA}{state-of-the-art }
\newcommand{\OSSSL}{one-shot semi-supervised learning }
\newcommand{\OSSSLno}{one-shot semi-supervised learning}
\newcommand{\SSL}{semi-supervised learning }
\newcommand{\SSLno}{semi-supervised learning}
\newcommand{\FM}{FixMatch }
\newcommand{\HP}{hyper-parameter }
\newcommand{\HPs}{hyper-parameters }

\def\cvprPaperID{3920} 
\def\confYear{CVPR 2021}

\begin{document}

\title{Building One-Shot Semi-supervised (BOSS) Learning up to Fully Supervised Performance}

\author{Leslie N. Smith\\
	US Naval Research Laboratory\\
	Washington, DC\\
{\tt\small leslie.smith@nrl.navy.mil}
\and
	Adam Conovaloff\\
NRC Postdoctoral Fellow\\
US Naval Research Laboratory\\
Washington, DC\\
{\tt\small adam.conovaloff.ctr@nrl.navy.mil}
}

\maketitle

\begin{abstract}
Reaching the performance of fully supervised learning with unlabeled data and only labeling one sample per class might be ideal for deep learning applications. We demonstrate for the first time the potential for building one-shot semi-supervised (BOSS) learning on Cifar-10 and SVHN up to attain test accuracies that are comparable to fully supervised learning.  Our method combines class prototype refining, class balancing, and self-training. A good prototype choice is essential and we propose a technique for obtaining iconic examples. In addition, we demonstrate that class balancing methods substantially improve accuracy results in semi-supervised learning to levels that allow self-training to reach the level of fully supervised learning performance.  Rigorous empirical evaluations provide evidence that labeling large datasets is not necessary for training deep neural networks.  
We made our code available at \url{https://github.com/lnsmith54/BOSS} to facilitate replication and for use with future real-world applications.
\end{abstract}

\section{Introduction}

In recent years, deep learning has achieved \SOTA performance for computer vision tasks such as image classification.
However, a major barrier to the wider-spread adoption of deep neural networks for new applications is that training \SOTA deep networks typically requires thousands to millions of labeled samples to perform at high levels of accuracy and to generalize well.

Unfortunately, manual labeling is labor intensive and might not be practical if labeling the data requires specialized expertise, such as in medical, defense, and scientific applications.
In typical real-world scenarios for deep learning, one often has access to large amounts of unlabeled data but lacks the time or expertise to label the required massive numbers needed for training, validation, and testing. 
An ideal solution might be to achieve performance levels that are equivalent to fully supervised trained networks with only one manually labeled image per class.

In this paper we investigate the potential for building one-shot semi-supervised (BOSS) learning up to achieve comparable performance as fully supervised training.
To date, \OSSSL has been little studied and viewed as difficult.
We build on the recent observation that \OSSSL is plagued by class imbalance problems \cite{smith2020empirical}.
In our context, class imbalance refers to a trained network with near 100\% accuracy on a subset of classes and poor performance on other classes. 
However, we are the first to apply data imbalance methods to unlabeled data.

Specifically, we demonstrate that good prototypes are crucial for successful \SSL and propose a prototype replacement method for the poorly performing classes.
Also, we make use of the \SOTA in \SSL methods (i.e., FixMatch \cite{sohn2020fixmatch}) in our experiments.
To combat class imbalance, we tested several variations of methods found in the literature for data imbalance problems \cite{johnson2019survey}, which refers to the situation where the number of training samples per class vary substantially.
We are the first to demonstrate that these methods significantly boost the performance of one-shot semi-supervised learning.
Combining these methods with self-training \cite{rosenberg2005semi} makes it possible  on Cifar-10 and SVHN to attain comparable performance as fully supervised trained deep networks.


Our contributions are:
\begin{enumerate}
	\item We rigorously demonstrate for the first time the potential for \OSSSL to reach test accuracies with Cifar-10 and SVHN that are comparable to fully supervised learning.
	\item We propose the concept of class balancing on unlabeled data and investigate their value of for \OSSSLno.  We introduce a novel measure of minority and majority classes and propose four class balancing methods  that improve the performance of \SSLno.
	\item We investigate the causes for poor performance and hyper-parameter sensitivity.  We hypothesis two causes and demonstrate solutions that improve performance.
\end{enumerate}

\section{Related Work}
\label{sec:related}

\textbf{Semi-supervised learning:} 
Semi-supervised learning is a hybrid between supervised and unsupervised learning, which combines the benefits of both and is better suited to real-world scenarios where unlabeled data is abundant.
As with supervised learning, \SSL defines a task (i.e., classification) from labeled data but typically it requires much fewer labeled samples.
In addition, semi-supervised learning leverages feature learning from unlabeled data to avoid overfitting the limited labeled samples.
Semi-supervised learning is a large and mature field and there are several surveys and books on semi-supervised learning methods \cite{zhu2005semi,van2020survey,chapelle2009semi,zhu2009introduction} for the interested reader.
In this Section we mention only the most relevant of recent methods.

Recently there have been a series of papers on \SSL from Google Reseach, including MixMatch \cite{berthelot2019mixmatch} , ReMixMatch \cite{berthelot2019remixmatch}, and FixMatch \cite{sohn2020fixmatch}.  
MixMatch combines consistency regularization with data augmentation \cite{sajjadi2016regularization}, entropy minimization (i.e., sharpening) \cite{grandvalet2005semi}, and mixup \cite{zhang2017mixup}.
ReMixMatch improved on MixMatch by incorporating distribution alignment and augmentation anchors.
Augmentation anchors are similar to pseudo-labeling.
FixMatch is the most recent and demonstrated \SOTA \SSL performance. 
In addition, the FixMatch paper has a discussion on \OSSSL with Cifar-10.

The FixMatch algorithm \cite{sohn2020fixmatch} is primarily a combination of consistency regularization \cite{sajjadi2016regularization,zhai2019s4l} and pseudo-labeling \cite{lee2013pseudo}.
Consistency regularization utilizes unlabeled data by relying on the assumption that the model should output the same predictions when fed perturbed versions as on the original image.
Consistency regularization has recently become a popular technique in unsupervised, self-supervised, and semi-supervised learning \cite{van2020survey,zhai2019s4l}.
Several researchers have observed that strong data augmentation should not be used when infering pseudo-labels for the unlabeled data but should be employed for consistency regularization \cite{sohn2020fixmatch,xie2019self}.
Pseudo-labeling is based on the idea that one can use the model to obtain artificial labels for unlabeled data by retaining pseudo-labels for samples whose probability are above a predefined threshold.

A recent survey of \SSL \cite{van2020survey} provides a taxonomy of classification algorithms.
One of the methods in \SSL is self-training iterations \cite{triguero2015self,rosenberg2005semi} where a classifier is iteratively trained on labeled data plus high confidence pseudo labeled data from previous iterations.
In our experiments we found that self-training provided a final boost to make the performance comparable to supervised training with the full labeled training dataset.

\textbf{Class imbalance:} 
Smith and Conovaloff \cite{smith2020empirical} demonstrated that in \OSSSL there are large variation in class performances, with some classes achieving near 100\% test accuracies while other classes near 0\% accuracies.
That is, strong classes starve the weak classes, which is analogous to the class imbalance problem \cite{johnson2019survey}.
This observation suggests an opportunity to improve the overall performance by actively improving the performance of the weak classes.

We borrowed techniques from the literature on training with imbalanced data \cite{johnson2019survey,wang2012multiclass,sun2007cost} (i.e., some classes having many more training samples than other classes) to experiment with several methods for improving the performance of the weak classes.
However, with unlabeled data, labels to define the ground truth as to minority and majority classes do not exist.
In this paper, we propose using the pseudo-labels as a surrogate to the ground truth for example class counting.
Our experiments demonstrate that combining the counting of the pseudo-labels and methods for handling data imbalance substantially improves performance.
Methods for handling class imbalance can be grouped into two categories: data-level and algorithm-level methods.
Data-level techniques \cite{wang2012multiclass} reduce the level of imbalance by undersampling the majority classes and oversampling the minority classes.
Algorithm-level techniques \cite{sun2007cost} are commonly implemented with smaller loss factor weights for the training samples belonging to the majority classes and larger weights for the training samples belonging to the minority classes.
In our experiments we tested variations of both types of methods and a hybrid of the two.


\textbf{Meta-learning:} 
Our scenario superficially bears similarity to few-shot meta learning \cite{koch2015siamese,vinyals2016matching,finn2017model,snell2017prototypical}, which is a highly active area of research.
The majority of the work in this area relies on a large labeled dataset with similar data statistics but this can be an onerous requirement for new applications.
While there are some recent efforts in unsupervised pretraining for few-shot meta learning \cite{hsu2018unsupervised,antoniou2019assume}, our experiments with these methods demonstrated their inability to adequately perform in one-shot learning to bootstrap our process.
Specifically, unsupervised one-shot learning with only five classes obtained a test accuracy of about 50\% on high confidence samples and the accuracy dropped sharply when increasing the number of classes.

\section{BOSS Methodology}
\label{sec:BOSS}

\subsection{FixMatch}
\label{sec:fixmatch}

Since we build on FixMatch \cite{sohn2020fixmatch}, we briefly describe the algorithm and adopt the formalism used in the original paper.
For an N-class classification problem, let us define $ \chi = \{ (x_b, y_b) : b \in (1, . . . , B)  \}  $ as a batch of B labeled examples, where $x_b$ are the training examples and $y_b$ are their labels. 
We also define $ \mathcal{U} = \{ u_b : b \in (1, . . . , \mu) \}$ as a batch of $\mu$ unlabeled examples where $\mu = r_u B $ and $r_u$ is a hyperparameter that determines the ratio of $\mathcal{U}$ to $\chi$. 
Let $p_m(y | x)$ be the predicted class distribution produced by the model for input $x_b$. 
We denote the cross-entropy between two probability distributions $p$ and $q$ as $H(p, q)$. 

The loss function for FixMatch consists two terms: a supervised loss $L_s$ applied to labeled data and an unsupervised loss $L_u$ for the unlabeled data. 
$L_s $ is the cross-entropy loss on weakly augmented labeled examples:
\begin{equation}
L_s = \frac{1}{B} \sum_{b=1}^{B} H(y_b, p_m (y|\alpha(x_b)))
\label{eqn:supervised}
\end{equation}
where $\alpha(x_b)$ represent weak data augmentation on labeled sample $x_b$.

For the unsupervised loss, the algorithm computes the label based on weakly augmented versions of the image as $q_b = p_m(y|\alpha(u_b))$.
It is essential that the label is computed on weakly augmented versions of the unlabeled training samples and not on strongly augmented versions.
The pseudo-label is computed as $\hat{q_b} = \arg\max(q_b)$ and the unlabeled loss is given as:
\begin{equation}
L_u = \frac{1}{\mu} \sum_{b=1}^{\mu} \mathbbm{1}  (max( q_b ) \geq \tau ) H(\hat{q_b}, p_m(y| \mathcal{A}(u_b)))
\label{eqn:Lu1}
\end{equation}
where $\mathcal{A}(u_b)$ represents applying strong augmentation to sample $u_b$ and $\tau$ is a scalar confidence threshold that is used to include only high confidence terms.
The total loss is given by $ L = L_s + \lambda_u L_u$ where $\lambda_u$ is a scalar hyper-parameter.
Additional details on the \FM algorithm are available in the paper \cite{sohn2020fixmatch}.

\subsection{Prototype refining}
\label{sec:refining}

Previous work by Sohn, \etal \cite{sohn2020fixmatch} on \OSSSL relied on the dataset labels to randomly choose an example for each class.
The authors demonstrated that the choice of these samples significantly affected the performance of their algorithm.
Specifically, they ordered the CIFAR-10 training data by how representative they were of their class by utilizing fully supervised trained models and found that using more prototypical examples achieved a median accuracy of 78\% while the use of poorly representative samples failed to converge at all.
The authors acknowledged that their method for finding prototypes was not practical.
In contrast, we now present a practical approach for choosing an iconic prototype for each class.

In real-world scenarios, one's data is initially all unlabeled but it is not overly burdensome for an expert to manually sift through some of their dataset to find one iconic example of each class.
In choosing iconic images of each class, the labeler's goal is to pick images that represent the class objects well, while minimizing the amount of background distractors in the image.
While the labeler is choosing the most iconic examples to be class prototypes for one-shot training of the network, it is beneficial to designate the less representative examples as part of a validation or test dataset.
In our own experiments with labeled datasets Cifar-10 and SVHN, we did not rely on the training labels but reviewed a small fraction of the training data to manually choose class prototypes.

In addition, we also propose a simple iterative technique for improving the choice of prototypes because good prototypes are important to good performance.
After choosing prototypes, the next step is to make a training run and examine the class accuracies.
For any class with poor accuracy relative to the other classes, it is likely that a better prototype can be chosen.  
We recommend returning to the unlabeled dataset to find replacement prototypes for only the poorly performing classes.
In our experiments we found doing this even once to be beneficial.

One might argue that prototype refining is as much work as labeling several examples per class and using many training samples will make it easier to train the model.
From only a practical perspective, labeling five or ten examples per class is not substantially more effort relative to labeling only one iconic example per class and prototype refining.
While in practice one may want to start with more than one example for ease of training, there are scientific, educational, and algorithmic benefits to studying \OSSSLno, which we discuss in our supplemental materials.
Also, non-representative examples can be included in a labeled test or validation dataset for use in evaluating the quality of the training.

\subsection{Class balancing}
\label{sec:balancing}

We believe a class imbalance problem is an important factor in training neural networks, not only in \OSSSL but also a factor for small to mid-sized datasets.
It is typical that a network with random weights usually outputs a single class label for every sample (i.e., randomly initialized networks do not generate random predictions). 
Hence, all networks start their training with elements of the class imbalance problem but the presence of large, balanced training data allows the network to overcome this problem.
Since class imbalance is always present when training deep networks, class balancing methods might always be valuable,  particularly when training on one-shot, few-shot, or small labeled datasets, and we leave further investigations of this for future work.


Unlike the data imbalance domain, the ground truth imbalance proportions are unknown with unlabeled datasets.
Our innovation here is to use the model generated pseudo-labels as a surrogate for class counting and estimating class imbalance ratios (i.e., determining majority and minority classes ).
Specifically, as the algorithm computes the pseudo-labels for all of the unlabeled training samples, it counts the number that fall within each class, which we designate as $\mathcal{C} = \{ c_n : n \in (1, . . . , N) \}$ where $N$ is the number of classes.
We assume a similar number of unlabeled samples in each class so the number of pseudo-labels in each class should also be similar.

Our first class balancing method is based on oversampling minority classes.
Our algorithm reduces the pseudo-labeling thresholds for minority classes to include more examples of the minority classes in the training.
Formally, in pseudo-labeling the following unsupervised loss function is used for the unlabeled data in place of Equation \ref{eqn:Lu1}:
\begin{equation}
L_u = \frac{1}{\mu} \sum_{b=1}^{\mu} \mathbbm{1} (max( q_b ) \geq \tau_n ) H(\hat{q_b}, q_b)
\label{eqn:pseudo1}
\end{equation}
where $q_b = p_m(y|\mathcal{A}(u_b))$, $\hat{q_b} = \arg\max(q_b)$, and $\tau_n$ is the class dependent threshold for inclusion in the unlabeled loss $L_u$.
We define the class dependent thresholds as:
\begin{equation}
\tau_n = \tau - \Delta (1 - \frac{c_n}{max(\mathcal{C})})
\label{eqn:pseudo2}
\end{equation}
where $c_n$ is the number of pseudo-labeled in class $n, max(\mathcal{C})$ is the maximum count of all the classes, and $\Delta$ is a scalar hyper-parameter ($\tau  > \Delta > 0$) guiding how much to lower the threshold for minority classes.
Hence, the most frequent class will use a threshold of $\tau$ while minority classes will use lower thresholds, down to $ \tau - \Delta $.

The next two class balancing methods are variations on loss function class weightings.
In the \FM algorithm, all unlabeled samples above the threshold are included in Equation \ref{eqn:pseudo1} with the same weight.
Instead, our second class balancing algorithm becomes:
\begin{equation}
L_u = \frac{1}{Z \mu} \sum_{b=1}^{\mu} \mathbbm{1} (max( q_b \geq \tau_n )) H(\hat{q_b}, q_b) / c_n
\label{eqn:pseudo3}
\end{equation}
where the loss terms are divided by $c_n$   and $ Z $ is a normalizing factor that makes $L_u$ the same magnitude as without this weighting scheme (this allows the unlabeled loss weighting $\lambda_u$ to remain the same).

Our third class balancing algorithm is identical to the previous method except it uses an alternate class count $\hat{c}_u$ in Equation \ref{eqn:pseudo2}.
We define $\hat{c}_u$ using only the high confidence pseudo-labeled samples (i.e., samples that are above the threshold).
The intuition of this third method is that each of the classes should contribute equally to the loss $L_u$ (i.e., each sample's loss is divided by the number of samples of that class included in  $L_u$).
In practice, this method's weights might be an order of magnitude larger than the previous method's weights, which might contribute to training instability, so we compare both methods in Section \ref{sec:expBalancing}.

Our fourth class balancing algorithm is a hybrid of the data and algorithmic methods.  Specifically, it is a  combination of our class balancing methods 1 and 3.  Our experiments with this hybrid method demonstrates the benefits of combining the class balancing methods.

\subsection{Self-training iterations}
\label{sec:iter}

Labeled and unlabeled data play different roles in \SSLno.  
Here we propose self-training iterations where the pseudo-labels of the highest confidence unlabeled training samples are combined with labeled samples in a new iteration.
Increasing the number of labeled samples per class improves performance, and substantially reduces training instability and performance variability.
Although some of these pseudo-labels might be wrong, we rely on the observation that the training of deep networks are robust to small amounts of labeling noise (i.e., labeling noise of less than 10\% does not harm the trained network's performance \cite{algan2019image}).
Hence, we aimed to achieve a $90\%$ accuracy from \SSL with the class balancing methods.

Self-training in BOSS adds to the testing stage a computation of the model predictions on all of the unlabeled training data.  
These are sorted from the the highest prediction probabilities down and the dataset is saved.
After the original training run, the labeled data can be combined with a number of the highest prediction samples from each class and a subsequent self-training iteration run can use the larger labeled dataset for retraining a new network.
We experimented with labeling 5, 10, 20, and 40 of the top predictions per class and the results are reported in Section \ref{sec:expIter}.


\begin{table*}
	\begin{center}
		\begin{tabular}{|c|c|c|c|c|c|c|c|c|c|c|c|}
			\hline
			Set & airplane & auto & bird & cat & deer & dog & frog & horse & ship & truck & Mean \\
			\hline\hline
			1 & $ 29 $  & $ 98 $  & $ 71 $  & $ 89 $  & $ 97 $  & $ 16 $  & $ 98 $  & $ 97 $  & $ 97 $  & $ 97 $ & $ 79 $ \\
			\hline
			2 & $ 28 $  & $ 99 $  & $ 70 $  & $ 43 $  & $ 97 $  & $ 89 $  & $ 98 $  & $ 97 $  & $ 98 $  & $ 0 $   & $ 72 $ \\
			\hline
			3 & $96 $  & $ 98 $  & $ 63 $  & $ 20 $  & $ 97 $  & $ 96 $  & $ 98 $  & $ 87 $  & $ 98 $  & $ 97 $   & $ 86 $ \\
			\hline
			4 & $ 29 $  & $ 98 $  & $ 65 $  & $ 10 $  & $ 96 $  & $ 32 $  & $ 98 $  & $ 97 $  & $ 97 $  & $ 96 $  & $ 72 $  \\
			\hline
			5 & $ 28 $  & $ 97 $  & $ 70 $  & $ 46 $  & $ 96 $  & $ 48 $  & $ 53 $  & $ 76 $  & $ 96 $  & $ 97 $  & $ 72 $  \\
			\hline
			6 & $ 80 $  & $ 98 $  & $ 71 $  & $ 52 $  & $ 97 $  & $ 92 $  & $ 98 $  & $ 87 $  & $ 98 $  & $ 97 $  & $ 82 $  \\
			\hline
			7 & $ 28 $  & $ 99 $  & $ 75 $  & $ 54 $  & $ 95 $  & $ 86 $  & $ 95 $  & $ 86 $  & $ 96 $  & $ 94 $  & $ 83 $  \\
			\hline		
		\end{tabular}
	\end{center}
	\caption{\textbf{Class accuracies.} One-shot semi-supervised average (of 2 runs) class accuracies for Cifar-10 test data with the \FM model, that was trained on sets of manually chosen prototypes for each class.  Prototype set 6 was modified from set 2 and prototype set 7 was modified from set 4 (i.e., prototype refining).
	}
	\label{tab:classAcc1}
\end{table*}


\section{Experiments}
\label{sec:exp}

In this Section we demonstrate that the BOSS algorithms can achieve comparable performance with fully-supervised training of Cifar-10 \cite{krizhevsky2009learning} and SVHN \cite{netzer2011reading}.
We compare our results to FixMatch\footnote{With appreciation, we acknowledge the use of the code kindly provided by the authors at \url{https://github.com/google-research/fixmatch} } \cite{sohn2020fixmatch} and demonstrate the value of our approach.
Our experiments use a Wide ResNet-28-2 \cite{zagoruyko2016wide} that matches the FixMatch reported results and we used the same cosine learning rate schedule described by Sohn, \etal \cite{sohn2020fixmatch}. 
We repeated our experiments with a ShakeNet model \cite{gastaldi2017shake} and obtained similar result that lead to the same insights and conclusions.
Our hyper-parameters were in a small range and the specifics are provided in the supplementary materials.
For data and data augmentation, we used the default augmentation in FixMatch but additional experiments (not shown) did show that using RandAugment \cite{cubuk2019randaugment} for strong data augmentation provides  a slight improvement.
Our runs with fully supervised learning of the Wide ResNet-28-2 model produced a test accuracy of $ 94.9 \pm 0.3 \% $ for Cifar-10 \cite{krizhevsky2009learning} and  test accuracy of $ 98.26 \pm 0.04 \% $ for SVHN \cite{netzer2011reading}, which we use for our basis of comparison.
We made our code available at \url{https://github.com/lnsmith54/BOSS} to facilitate replication and for use with future real-world applications.


\subsection{Choosing prototypes and prototype refining}
\label{sec:expPrto}

For our experiments with Cifar-10, we manually reviewed the first few hundred images and choose five sets of prototypes that we will refer to as class prototype sets 1 to 5.
However, the practioner need only create one set of class prototypes and can perform prototype refining, as we describe below.

Table \ref{tab:classAcc1} presents the averaged (over two runs) test accuracies for each class, computed from FixMatch on the Cifar-10 test dataset for each of the prototype sets 1 to 5.
This Table illustrates that a good choice of prototypes (i.e., set=3) can lead to good performance in most of the classes, which enables a good overall performance.
Table \ref{tab:classAcc1} also shows that for other sets the class accuracies can be quite high for some classes while low for other classes.
Hence, the poor performance of some classes implies that the choice of prototypes for these classes in those sets can be improved.
In prototype refining, one simply reviews the class accuracies to find which prototypes should be replaced.

We demonstrate prototype refining with two examples.  
The airplane and truck class accuracies in set 2 are poor so we replaced these two prototypes and name this set 6.
In set 4, the cat and dog classes are performing poorly so we replaced these two prototypes and name this set 7.
Table \ref{tab:classAcc1} shows the class accuracies for sets 6 and 7 and these results are better than the original sets; that is, prototype refining of these two sets raised the overall test accuracies from 72\% up to 82-83\%.



\begin{table*}
	\begin{center}
		\begin{tabular}{|c|c|c|c|c|c|c|c|c|c|}
			\hline
			&  & \multicolumn{4}{|c|}{BOSS balance method} & \multicolumn{4}{|c|}{Self-training}    \\
			\hline
			Set  & FixMatch & 1 & 2 & 3 & 4 & +5 & +10 & +20 & +40    \\
			
			\hline\hline
			1 & $ 79 \pm 1 $  & $ \mathbf{91.4 \pm 2} $  & $ 90 \pm 5 $  & $  84 \pm 6 $ & $ 88 \pm 2 $ & $ 94.8 $ & $ 95.2 $ & $ 95.2 $ & $ 95.2 $   \\
			& $ [61]  $  & $ [75] $  & $ [85] $  & $ [70] $  & $ [67] $ & $ \pm 0.1 $ & $ \pm 0.1 $ & $ \pm 0.1 $ & $ \pm 0.1 $   \\
			\hline
			2 & $ 74 \pm 5 $  & $ \mathbf{91.8 \pm 1} $  & $ 90 \pm 3 $  & $  88 \pm 2 $ & $ 80 \pm 14 $ & $ 93.6 $ & $ 95.1 $ & $ 95.1 $ & $ 95.1 $   \\
			& $ [58]  $  & $ [85] $  & $ [83] $  & $ [81] $  & $ [89] $ & $ \pm 0.2  $ & $ \pm 0.1 $ & $ \pm 0.3 $ & $ \pm 0.2 $   \\
			\hline
			3 & $ 86 \pm 1 $  & $ 92.8 \pm .2 $  & $ 91 \pm 2 $  & $ 91 \pm 3 $ & $ \mathbf{92.8 \pm .1} $ & $ 94.6 $ & $ 94.8 $ & $ 94.9 $ & $ 95.2 $   \\
			& $ [88]  $  & $ [93] $  & $ [91] $  & $ [89] $  & $ [87] $ & $ \pm 0.5 $ & $ \pm 0.5 $ & $ \pm 0.1  $ & $ \pm 0.1 $   \\
			\hline
			4 & $ 74 \pm 8 $  & $ 77.7 \pm .3 $  & $ 81 \pm 6 $  & $ 81 \pm 8 $ & $ \mathbf{90 \pm 7} $ & $ 94.9 $ & $ 94.9 $ & $ 94.9 $ & $ 95.1 $   \\
			& $ [73]  $  & $ [89] $  & $ [72] $  & $ [86] $  & $ [82] $ & $ \pm 0.1 $ & $ \pm 0.4 $ & $ \pm 0.5  $ & $ \pm 0.3 $   \\
			\hline
			5 & $ 69 \pm 7 $  & $ 86 \pm 7 $  & $ 89 \pm 6 $  & $ 83 \pm 10 $ & $ \mathbf{90 \pm 3} $ & $ 89.6 $ & $ 95.2 $ & $ 95.2 $ & $ 95.2 $   \\
			& $ [69]  $  & $ [86] $  & $ [73] $  & $ [87] $  & $ [85] $ & $ \pm 0.3 $ & $ \pm 0.1 $ & $ \pm 0.2 $ & $  \pm 0.1 $   \\
			\hline
			6 & $ 82 \pm 0.6 $  & $ 91.5 \pm 1 $  & $ 92 \pm .7 $  & $ 91.8 \pm 1 $ & $ \mathbf{92 \pm 1} $ & $ 94.6 $ & $ 95.1 $ & $ 94.7 $ & $ 94.9 $   \\
			& $ [87]  $  & $ [83] $  & $ [81] $  & $ [75] $  & $ [70] $ & $ \pm 0.1 $ & $ \pm 0.2 $ & $ \pm 0.1 $ & $ \pm 0.1 $   \\
			\hline			
			7 & $ 78 \pm 0.1 $  & $ 91.7 \pm .3 $  & $ 92.3 \pm .8 $  & $ 91.1 \pm 2.5 $ & $ \mathbf{93 \pm .3} $ & $ 94.9 $ & $ 94.7 $ & $ 94.9 $ & $ 95.1 $   \\
			& $ [56]  $  & $ [68] $  & $ [79] $  & $ [62] $  & $ [66] $ & $ \pm 0.1 $ & $ \pm 0.2 $ & $ \pm 0.1 $ & $ \pm 0.1 $   \\
			\hline			
			
		\end{tabular}
	\end{center}
	\caption{ \textbf{Main results.}  BOSS methods are compared using five sets of class prototypes (i.e., 1 prototype per class) for Cifar-10, plus two sets from prototype refining.  The FixMatch column shows test accuracies (average and standard deviation of 4 runs) for the original FixMatch code on the prototype sets.  The next four columns gives the accuracy results for the class balance methods (see text for a description of class balance methods).  Results for the PyTorch reimplementation of FixMatch and modified with the BOSS methods are shown in brackets [.].  The self-training iteration was performed with the top pseudo-labels from the run shown in bold and the results are in the next four columns. 
	}
	\label{tab:balanceAcc1}
\end{table*}


\subsection{Class balancing}
\label{sec:expBalancing}


In this Section we report the results from FixMatch and demonstrate substantial improvements with the class balancing methods in BOSS.
Table \ref{tab:balanceAcc1} presents our main results, which illustrates the benefits from prototype refining, class balancing, and one self-training iteration.
The first five rows in the table list the results for the five sets of class prototypes (i.e., 1 prototype per class) for Cifar-10.
Rows for sets 6 and 7 provides the results for prototype refining of the original sets 2 and 4, respectively.  
The FixMatch column shows results (i.e., average and standard deviation over four runs) for the original FixMatch code on the prototype sets. 
The number within brackets [.] are results from a PyTorch reimplementation of FixMatch, that we discuss below.

The next four columns presents the BOSS results with class balancing methods.
As described in Section \ref{sec:balancing}, class balance method 1 represents oversampling of minority classes, balance methods 2 and 3 are two forms of class-based loss weightings, and balance = 4 is a  hybrid that combines balance methods 1 and 3.
The use of class balancing significantly improves on the original FixMatch results, with increases of up to 20 absolute percentage points.
Generally, the hybrid class balance method 4 is best, except when instabilities hurt the performance.
The performance is generally in the 90\% range with good performance across all the classes, which enables the self-training iteration to bump the accuracies to be comparable to the test accuracy from supervised training on the full labeled training dataset.

Table \ref{tab:balanceAcc1} indicates that good class prototypes (i.e., sets 3, 6, and 7) result in test accuracies near 90\% and low variance between runs.
However, when some of the class prototypes are inferior, some of the of the training runs exhibit instabilities that cause lower averaged accuracies and higher variance.
We provide a discussion in Section \ref{sec:expHPsensitivity} on the cause of these instabilities and on how to improve these results.

\textbf{PyTorch version:}
We have taken advantage of a PyTorch reimplementation\footnote{With appreciation, we acknowledge the use of the code provided at \url{https://github.com/CoinCheung/fixmatch} } of the original TensorFlow version of the \FM code to test our proposed BOSS methods in PyTorch.
Table \ref{tab:balanceAcc1} reports the best test accuracies for the PyTorch version in the brackets [.].

It is clear to us that the researcher who reimplemented FixMatch in PyTorch took care to replicate FixMatch.
In training with 4 labeled samples per class, his code obtained a test accuracy of $ 89 \pm 5\%$ for Cifar-10, compared to results of $ 87 \pm 3\% $ reported in the paper.
However, it is also clear from our experiments and Table \ref{tab:balanceAcc1} that there are substantial differences between the TensorFlow and PyTorch versions when comparing \OSSSLno.
We suggest that the sensitivity of \OSSSL reveals even minor differences that are invisible in fully supervised learning.

The PyTorch implementation results shown in Table \ref{tab:balanceAcc1} also shows that the class balancing methods improve the test accuracy over FixMatch.
In particular, class balance method 1 (i.e., oversampling) appears to improve the test accuracy more than the other methods.

\subsection{Self-training iterations}
\label{sec:expIter}

The final four columns of Table \ref{tab:balanceAcc1} list the results of performing one self-training iteration.
The self-training was initialized with the  original single labeled sample per class, plus  the most confident pseudo-labeled examples from the BOSS training run that is highlighted in bold.
For example, the `+5' columns means that five pseudo-labeled examples per class were combined with the original labeled prototypes to make a set with a total of 60 labeled examples.
These self-training results demonstrate that \OSSSL can reach comparable performance to the results from fully supervised training (i.e., 94.9\%), often with adding as few as 5 samples per class.
However, we expect that in practice, self-training by adding more samples per class will prove more reliable.


\begin{table*}
	\begin{center}
		\begin{tabular}{|c|c|c|c|c|c|c|c|c|c|}
			\hline
			&  & \multicolumn{4}{|c|}{BOSS balance method} & \multicolumn{4}{|c|}{self-training}    \\
			\hline
			set  & FixMatch & 1 & 2 & 3 & 4 & +5 & +10 & +20 & +40    \\
			\hline\hline
			1 & $ 95.9 \pm 3 $  &  $ \mathbf{97.4 \pm .2} $  & $ 96.4 \pm .9 $  & $ 95.7 \pm 1.6 $ & $ 96.8 \pm .1 $ & $ 97.9 $ & $ 97.9 $ & $ 97.9 $ & $ 97.8 $   \\
			\hline
			2 & $ 91.5 \pm 3 $  &  $ \mathbf{97.4 \pm .1} $  & $ 97.1 \pm .1 $  & $ 97.1 \pm .1 $ & $ 95.6 \pm .1 $ & $ 94.1 $ & $ 97.9 $ & $ 97.6 $ & $ 97.7 $   \\
			\hline
			3 & $ 93.9 \pm .1 $  &  $ \mathbf{97.3 \pm .3} $  & $ 97.2 \pm .2 $  & $ 92 \pm 7 $ & $ 91.3 \pm .3 $ & $ 97.8 $ & $ 97.9 $ & $ 97.8 $ & $ 97.9 $   \\
			\hline
			4 & $ 89.2 \pm 12 $  &  $ \mathbf{96.5 \pm .6} $  & $ 90 \pm 10 $  & $ 89 \pm 11 $ & $ 83 \pm 16 $ & $ 97.6 $ & $ 96.7 $ & $ 97.0 $ & $ 98.0 $   \\
			\hline
			
		\end{tabular}
	\end{center}
	\caption{\textbf{SVHN.} BOSS methods are compared using four sets of class prototypes (i.e., 1 prototype per class) for SVHN.  The FixMatch column shows results for the original FixMatch code on the prototype sets. The next four columns gives the accuracy results for the class balance methods  Results are an average of test accuracies for four runs.  The self-training iteration was performed on the results from the class balancing shown in bold.
	}
	\label{tab:balanceAccSVHN}
\end{table*}

\subsection{SVHN}
\label{sec:svhn}

SVHN is obtained from house numbers in Google Street View images and is used for recognizing digits (i.e., 0 -- 9) in natural scene images. 
Visual review of the images show that the training samples are of poor quality (i.e., blurry) and often contain distractors (i.e.,  multiple digits in an image).
Because of the quality issue, we needed to review several hundred unlabled training samples in order to find four class prototype sets that are reported in Table \ref{tab:balanceAccSVHN}.

Even though the SVHN training images are of poorer quality than the Cifar-10 training images, \OSSSL with FixMatch on sets of prototypes produced higher test accuracies than with Cifar-10.
Table \ref{tab:balanceAccSVHN} presents equivalent results for the SVHN dataset as those results that were reported in Table \ref{tab:balanceAcc1} for Cifar-10.
Since the results for FixMatch are all above 89\%, we did not perform prototype refining on any of these sets.
However, here too the class balancing methods increase the test accuracies above the FixMatch results.
With these four class prototype sets, class balance method 1 produces the best results.
The test accuracies from balance method 1 are approximately 1\% lower than the fully supervised results of $ 98.26 \pm 0.04 \%. $
The improvements from self-training were small and the best results fell about 0.5\% below the results of of fully supervised training.
We believe the differences between Cifar-10 and SVHN are related to the natures of the datasets.

\begin{figure}[t]
	\begin{center}
		\includegraphics[width=0.9\linewidth]{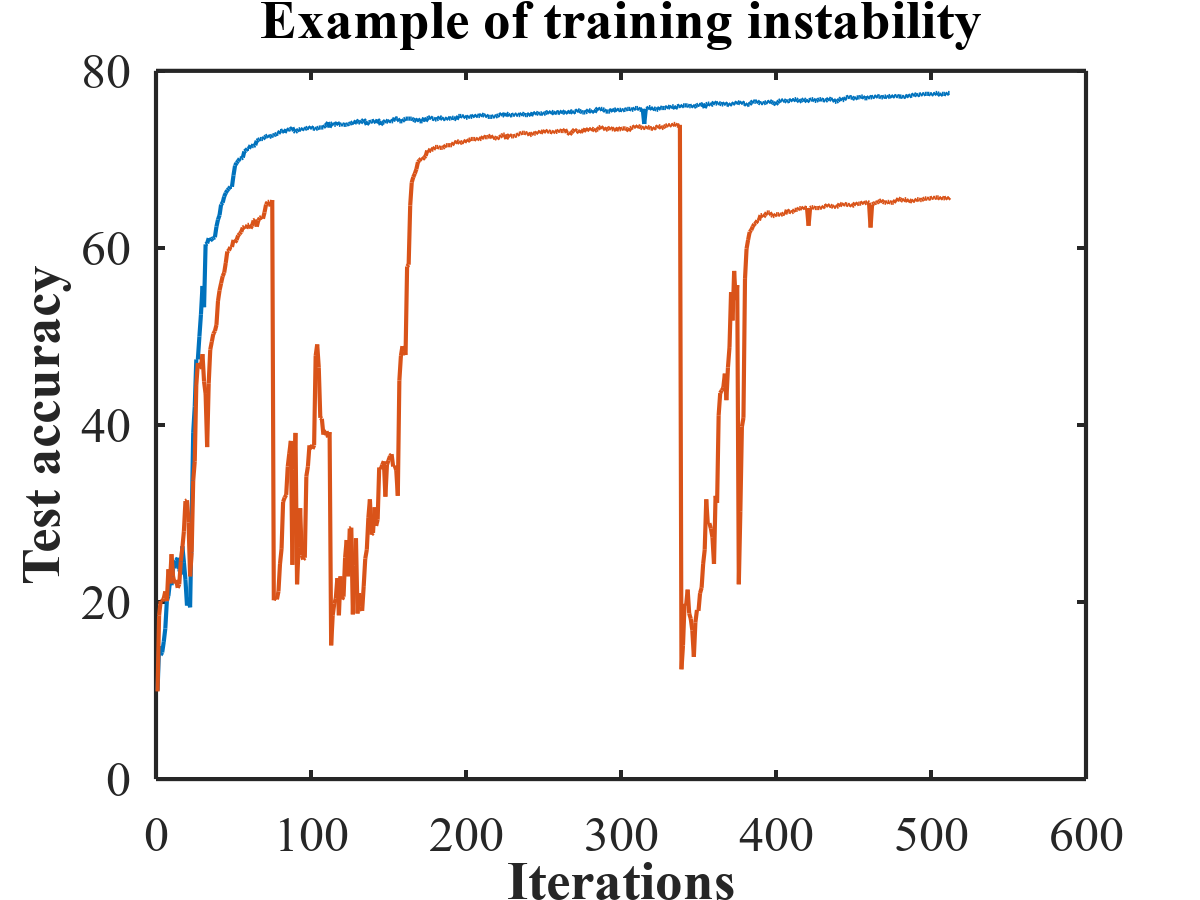}
	\end{center}
	\caption{An example of training to a poor local minimum (blue) and training with instabilities (red). Both end with poor test accuracies but for different reasons. }
	\label{fig:instability}
\end{figure}

\begin{table*}
	\begin{center}
		\begin{tabular}{|c|c|c|c|c|c|c|c|c|}
			\hline
			Set  & Balance & Description & WD & LR & $ \Delta $ & $ \lambda_u $ & $ \tau $ & Accuracy (\%)    \\
			\hline\hline
			1 & 3  & Instabilities  & $ 8 \times 10^{-4} $ & 0.06 &  0  &  1 & 0.9   & $ 84 \pm 6  $   \\
			\hline
			1 & 3  & Decrease $ \lambda_u $  & $ 8 \times 10^{-4} $ & 0.06 &  0  & 0.5 & 0.9   & $ 87 \pm 1  $   \\
			\hline\hline
			2 & 4  & Instabilities  & $ 8 \times 10^{-4} $ & 0.06 &  0.25  &  1 & 0.95   & $ 80 \pm 14  $   \\
			\hline
			2 & 4  & Decrease $ \Delta$, WD, LR  & $ 6 \times 10^{-4} $ & 0.04 &  0.1  & 1 & 0.95   & $ 94.5 \pm 0.1  $   \\
			\hline\hline
			4 & 1  & Local min  & $ 8 \times 10^{-4} $ & 0.06 &  0.25  &  1 & 0.9   & $ 77.5 \pm 0.1  $   \\
			\hline
			4 & 1  & Increase $ \Delta, \tau $  & $ 8 \times 10^{-4} $ & 0.06 &  0.3  &  1 & 0.95   & $ 93.2 \pm 0.2  $   \\
			\hline\hline
			4 & 2  & Local min  & $ 8 \times 10^{-4} $ & 0.06 &  0  &  1 & 0.9   & $ 81 \pm 6  $   \\
			\hline
			4 & 2  & Increase $ \lambda_u $  & $ 8 \times 10^{-4} $ & 0.06 &  0  &  2 & 0.9   & $ 92 \pm 2  $   \\
			\hline\hline
			4 & 3  & Local min  & $ 8 \times 10^{-4} $ & 0.06 &  0  &  1 & 0.9   & $ 81 \pm 8  $   \\
			\hline
			4 & 3  & Increase $ \lambda_u $  & $ 8 \times 10^{-4} $ & 0.06 &  0  &  2 & 0.9   & $ 88 \pm 3  $   \\
			\hline\hline
			5 & 1  & Instabilities  & $ 8 \times 10^{-4} $ & 0.06 &  0.25  &  1 & 0.95   & $ 86 \pm 7  $   \\
			\hline
			5 & 1  & Decrease $ \Delta $  & $ 8 \times 10^{-4} $ & 0.06 &  0.1  &  1 & 0.95   & $ 90.7 \pm 0.1  $   \\
			\hline\hline
			5 & 2  & Instabilities  & $ 8 \times 10^{-4} $ & 0.06 &  0  &  1 & 0.9   & $ 89 \pm 6  $   \\
			\hline
			5 & 2  & Decrease $ \lambda_u $  & $ 8 \times 10^{-4} $ & 0.06 &  0  &  0.75 & 0.9   & $ 91.7 \pm 1  $   \\
			\hline\hline
			5 & 3  & Instabilities  & $ 8 \times 10^{-4} $ & 0.06 &  0  &  1 & 0.9   & $ 83 \pm 10  $   \\
			\hline
			5 & 3  & Decrease WD, LR  & $ 6 \times 10^{-4} $ & 0.04 &  0  &  1 & 0.9   & $ 93.5 \pm 2  $   \\
			\hline
		\end{tabular}
	\end{center}
	\caption{Illustration of the sensitivity to the \HPs WD, LR, $ \Delta$, $ \lambda_u $ and $ \tau $.  See the text for guidance on how to tune these hyper-parameters for situations with inferior performance due to instabilities or local minimums. 
	}
	\label{tab:tuneHP}
\end{table*}


\subsection{Investigation of training instabilities}
\label{sec:expHPsensitivity}

In our experiments we observed high sensitivity of \OSSSL performance to the choices for the hyper-parameters and the class prototypes sets.
That is, we observed that good choices for the prototypes and prototype refining significantly reduced the instabilities and the variability of the results (i.e., few instabilities were encountered for Cifar-10 prototype sets 3, 6, and 7 so the final accuracies were higher and the standard deviations of the results were lower).
In sets where the performance was inferior, there was always at least one class that performed poorly.
In addition, we found that the \HP values made a significant difference.


We investigated the cases of poor performance and discovered that there were two different situations.
Figure \ref{fig:instability} provides examples of test accuracies during the training for both situations.
The blue curve is the test accuracy where in one training run the network learns a final test accuracy of 77\% 
We hypothesize that in this situation the network can get stuck in a poor local minimum that is due to poor prototype choices and can be improved with prototype refining or by \HP fine tuning.
The red curve in Figure \ref{fig:instability} is an example of the other case and here the training is dominated by instabilities (i.e., where the model suddenly diverges during training) and the final test accuracy is 65\%.
Interestingly, we found that it is important when tuning the \HPs to identify which scenario is occurring.

Our experiments with training instabilities (i.e., the red curve) implied that they can be caused by too much class balancing.
We hypothesize that when the model struggles to classify some of the classes, the class balancing methods can force the pseudo-labeling to mislabel samples in order to have the appearance of class balance.
In these cases, it is better to reduce the amount of class balancing by using a smaller value for $ \Delta $ for class balance methods 1 and 4, and using a smaller value for $ \lambda_u $ for class balance methods 2 and 3.
In addition, we observed that decreasing weight decay (WD) and the learning rate (LR) improves performance when there were instabilities.

On the other hand, if the inferior performance is due to poor local minimum (i.e., the blue curve), one can either improve the class prototypes (i.e., prototype refining) or increase the amount of class balancing.
This is the opposite of what should do for instabilities; that is, one can use a larger value for $ \Delta $ for class balance methods 1 and 4, use a larger value for $ \lambda_u $ for class balance methods 2 and 3, or increase weight decay (WD) and the learning rate (LR).
We also observed that it helps to increase $ \tau $ if there are instabilities and to decrease $ \tau $ in the poor local minimum situation.

Table \ref{tab:tuneHP} demonstrates how to improve the results presented in Table \ref{tab:balanceAcc1} (for consistency we used the same \HP values for all of the class balance runs shown in Table \ref{tab:balanceAcc1}).
Table \ref{tab:tuneHP} contains results of \HP fine tuning where we reported earlier test accuracies below 85\%.
We list the class prototype set (Set), the BOSS class balancing method (Balance), weight decay (WD), initial learning rate (LR), the change in the confidence threshold for minority classes ($\Delta$), the unlabeled loss multiplicative factor ($\lambda_u$), the confidence threshold ($\tau$), and the final test accuracy.
Furthermore, we provide a short description that indicates if the training curve displays instabilities (i.e., the red curve in Figure \ref{fig:instability}) or a poor local minimum (i.e., the blue curve).
Or the description points out the \HPs that were tuned to improve the performance.


The examples in Table \ref{tab:tuneHP} show improved results for both the problem of instability and for poor local minimums.
The examples include modifying $\Delta$, weight decay, learning rate, and $\tau$.
In most cases the final accuracies are improved substantially with small changes in the \HP values.
This demonstrates the sensitivity of \OSSSL to \HP values.

While this sensitivity can be challenging in practice, we note that this sensitivity can also lead to new opportunities.
For example, often researchers propose new network architectures, loss functions, and optimization functions that are tested in the fully supervised regime where small performance gains are used to claim a new state-of-the-art.
If these algorithms were instead tested in \OSSSLno, more substantial differences in performance would better differentiate methods.
Along these lines, we also advocate the use of \OSSSL with AutoML and neural architecture search (NAS) \cite{elsken2018neural} to find optimal hyper-parameters and architectures.

\section{Conclusions}
\label{sec:conclusions}

The BOSS methodology relies on simple concepts: choosing iconic training samples with minimal background distractors, employing class balancing techniques, and self-training with the highest confidence pseudo-labeled samples.
Our experiments in Section \ref{sec:exp} demonstrate the potential of training a network with only one sample per class and we have confirmed the importance of class balancing methods.
While our methods have limitations (as discussed in the supplemental materials), this paper breaks new ground in one-shot semi-supervised learning and attains high performance.
BOSS brings one-shot and few-shot \SSL closer to reality. 

We proposed the novel concept of class balancing on unlabeled data.  We introduced a novel way to measure  class imbalance with unlabeled data and proposed four class balancing methods that improve the performance of \SSLno.  In addition, we investigated hyper-parameter sensitivity and the causes for weak performance (i.e., training instabilities), where we proposed two opposite sets of solutions.

Our work provides researchers with the following observations and insights:
\begin{enumerate}
	\item There is evidence that labeling a large number of samples might not be necessary for training deep neural networks to high levels of performance.
	\item All networks have a class imbalance problem to some degree. Examining class accuracies relative to each other provides insights into the network's training.
	\item Each training sample can affect the training.  One-shot \SSL provides a mechanism to study the atomic impact of a single sample.  This opens up the opportunity to investigate the factors in a sample that help or hurt training performance.
\end{enumerate}


{\small
\bibliographystyle{ieee_fullname}
\bibliography{boss}

\begin{thebibliography}{10}\itemsep=-1pt

\bibitem{algan2019image}
G{\"o}rkem Algan and Ilkay Ulusoy.
\newblock Image classification with deep learning in the presence of noisy
  labels: A survey.
\newblock {\em arXiv preprint arXiv:1912.05170}, 2019.

\bibitem{antoniou2019assume}
Antreas Antoniou and Amos Storkey.
\newblock Assume, augment and learn: Unsupervised few-shot meta-learning via
  random labels and data augmentation.
\newblock {\em arXiv preprint arXiv:1902.09884}, 2019.

\bibitem{berthelot2019remixmatch}
David Berthelot, Nicholas Carlini, Ekin~D Cubuk, Alex Kurakin, Kihyuk Sohn, Han
  Zhang, and Colin Raffel.
\newblock Remixmatch: Semi-supervised learning with distribution alignment and
  augmentation anchoring.
\newblock {\em arXiv preprint arXiv:1911.09785}, 2019.

\bibitem{berthelot2019mixmatch}
David Berthelot, Nicholas Carlini, Ian Goodfellow, Nicolas Papernot, Avital
  Oliver, and Colin~A Raffel.
\newblock Mixmatch: A holistic approach to semi-supervised learning.
\newblock In {\em Advances in Neural Information Processing Systems}, pages
  5050--5060, 2019.

\bibitem{chapelle2009semi}
Olivier Chapelle, Bernhard Scholkopf, and Alexander Zien.
\newblock Semi-supervised learning (chapelle, o. et al., eds.; 2006)[book
  reviews].
\newblock {\em IEEE Transactions on Neural Networks}, 20(3):542--542, 2009.

\bibitem{cubuk2019randaugment}
Ekin~D Cubuk, Barret Zoph, Jonathon Shlens, and Quoc~V Le.
\newblock Randaugment: Practical data augmentation with no separate search.
\newblock {\em arXiv preprint arXiv:1909.13719}, 2019.

\bibitem{elsken2018neural}
Thomas Elsken, Jan~Hendrik Metzen, and Frank Hutter.
\newblock Neural architecture search: A survey.
\newblock {\em arXiv preprint arXiv:1808.05377}, 2018.

\bibitem{finn2017model}
Chelsea Finn, Pieter Abbeel, and Sergey Levine.
\newblock Model-agnostic meta-learning for fast adaptation of deep networks.
\newblock In {\em Proceedings of the 34th International Conference on Machine
  Learning-Volume 70}, pages 1126--1135. JMLR. org, 2017.

\bibitem{gastaldi2017shake}
Xavier Gastaldi.
\newblock Shake-shake regularization.
\newblock {\em arXiv preprint arXiv:1705.07485}, 2017.

\bibitem{grandvalet2005semi}
Yves Grandvalet and Yoshua Bengio.
\newblock Semi-supervised learning by entropy minimization.
\newblock In {\em Advances in neural information processing systems}, pages
  529--536, 2005.

\bibitem{hsu2018unsupervised}
Kyle Hsu, Sergey Levine, and Chelsea Finn.
\newblock Unsupervised learning via meta-learning.
\newblock {\em arXiv preprint arXiv:1810.02334}, 2018.

\bibitem{johnson2019survey}
Justin~M Johnson and Taghi~M Khoshgoftaar.
\newblock Survey on deep learning with class imbalance.
\newblock {\em Journal of Big Data}, 6(1):27, 2019.

\bibitem{koch2015siamese}
Gregory Koch, Richard Zemel, and Ruslan Salakhutdinov.
\newblock Siamese neural networks for one-shot image recognition.
\newblock In {\em ICML deep learning workshop}, volume~2. Lille, 2015.

\bibitem{krizhevsky2009learning}
Alex Krizhevsky, Geoffrey Hinton, et~al.
\newblock Learning multiple layers of features from tiny images.
\newblock 2009.

\bibitem{laine2016temporal}
Samuli Laine and Timo Aila.
\newblock Temporal ensembling for semi-supervised learning.
\newblock In {\em Fifth International Conference on Learning Representations},
  2017.

\bibitem{lee2013pseudo}
Dong-Hyun Lee.
\newblock Pseudo-label: The simple and efficient semi-supervised learning
  method for deep neural networks.
\newblock In {\em Workshop on challenges in representation learning, ICML},
  volume~3, page~2, 2013.

\bibitem{miyato2018virtual}
Takeru Miyato, Shin-ichi Maeda, Shin Ishii, and Masanori Koyama.
\newblock Virtual adversarial training: a regularization method for supervised
  and semi-supervised learning.
\newblock {\em IEEE transactions on pattern analysis and machine intelligence},
  2018.

\bibitem{netzer2011reading}
Yuval Netzer, Tao Wang, Adam Coates, Alessandro Bissacco, Bo Wu, and Andrew~Y
  Ng.
\newblock Reading digits in natural images with unsupervised feature learning.
\newblock 2011.

\bibitem{rosenberg2005semi}
Chuck Rosenberg, Martial Hebert, and Henry Schneiderman.
\newblock Semi-supervised self-training of object detection models.
\newblock {\em WACV/MOTION}, 2, 2005.

\bibitem{sajjadi2016regularization}
Mehdi Sajjadi, Mehran Javanmardi, and Tolga Tasdizen.
\newblock Regularization with stochastic transformations and perturbations for
  deep semi-supervised learning.
\newblock In {\em Advances in neural information processing systems}, pages
  1163--1171, 2016.

\bibitem{smith2020empirical}
Leslie~N Smith and Adam Conovaloff.
\newblock Empirical perspectives on one-shot semi-supervised learning.
\newblock {\em arXiv preprint arXiv:2004.04141}, 2020.

\bibitem{snell2017prototypical}
Jake Snell, Kevin Swersky, and Richard Zemel.
\newblock Prototypical networks for few-shot learning.
\newblock In {\em Advances in neural information processing systems}, pages
  4077--4087, 2017.

\bibitem{sohn2020fixmatch}
Kihyuk Sohn, David Berthelot, Chun-Liang Li, Zizhao Zhang, Nicholas Carlini,
  Ekin~D Cubuk, Alex Kurakin, Han Zhang, and Colin Raffel.
\newblock Fixmatch: Simplifying semi-supervised learning with consistency and
  confidence.
\newblock {\em arXiv preprint arXiv:2001.07685}, 2020.

\bibitem{sun2007cost}
Yanmin Sun, Mohamed~S Kamel, Andrew~KC Wong, and Yang Wang.
\newblock Cost-sensitive boosting for classification of imbalanced data.
\newblock {\em Pattern Recognition}, 40(12):3358--3378, 2007.

\bibitem{tarvainen2017mean}
Antti Tarvainen and Harri Valpola.
\newblock Mean teachers are better role models: Weight-averaged consistency
  targets improve semi-supervised deep learning results.
\newblock In {\em Advances in neural information processing systems}, 2017.

\bibitem{triguero2015self}
Isaac Triguero, Salvador Garc{\'\i}a, and Francisco Herrera.
\newblock Self-labeled techniques for semi-supervised learning: taxonomy,
  software and empirical study.
\newblock {\em Knowledge and Information systems}, 42(2):245--284, 2015.

\bibitem{van2020survey}
Jesper~E Van~Engelen and Holger~H Hoos.
\newblock A survey on semi-supervised learning.
\newblock {\em Machine Learning}, 109(2):373--440, 2020.

\bibitem{verma2019interpolation}
Vikas Verma, Alex Lamb, Juho Kannala, Yoshua Bengio, and David Lopez-Paz.
\newblock Interpolation consistency training for semi-supervised learning.
\newblock {\em arXiv preprint arXiv:1903.03825}, 2019.

\bibitem{vinyals2016matching}
Oriol Vinyals, Charles Blundell, Timothy Lillicrap, Daan Wierstra, et~al.
\newblock Matching networks for one shot learning.
\newblock In {\em Advances in neural information processing systems}, pages
  3630--3638, 2016.

\bibitem{wang2012multiclass}
Shuo Wang and Xin Yao.
\newblock Multiclass imbalance problems: Analysis and potential solutions.
\newblock {\em IEEE Transactions on Systems, Man, and Cybernetics, Part B
  (Cybernetics)}, 42(4):1119--1130, 2012.

\bibitem{xie2019unsupervised}
Qizhe Xie, Zihang Dai, Eduard Hovy, Minh-Thang Luong, and Quoc~V Le.
\newblock Unsupervised data augmentation for consistency training.
\newblock 2019.

\bibitem{xie2019self}
Qizhe Xie, Eduard Hovy, Minh-Thang Luong, and Quoc~V Le.
\newblock Self-training with noisy student improves imagenet classification.
\newblock {\em arXiv preprint arXiv:1911.04252}, 2019.

\bibitem{zagoruyko2016wide}
Sergey Zagoruyko and Nikos Komodakis.
\newblock Wide residual networks.
\newblock {\em arXiv preprint arXiv:1605.07146}, 2016.

\bibitem{zhai2019s4l}
Xiaohua Zhai, Avital Oliver, Alexander Kolesnikov, and Lucas Beyer.
\newblock S4l: Self-supervised semi-supervised learning.
\newblock In {\em Proceedings of the IEEE international conference on computer
  vision}, pages 1476--1485, 2019.

\bibitem{zhang2017mixup}
Hongyi Zhang, Moustapha Cisse, Yann~N Dauphin, and David Lopez-Paz.
\newblock mixup: Beyond empirical risk minimization.
\newblock {\em arXiv preprint arXiv:1710.09412}, 2017.

\bibitem{zhu2009introduction}
Xiaojin Zhu and Andrew~B Goldberg.
\newblock Introduction to semi-supervised learning.
\newblock {\em Synthesis lectures on artificial intelligence and machine
  learning}, 3(1):1--130, 2009.

\bibitem{zhu2005semi}
Xiaojin~Jerry Zhu.
\newblock Semi-supervised learning literature survey.
\newblock Technical report, University of Wisconsin-Madison Department of
  Computer Sciences, 2005.

\end{thebibliography}
}

\appendix

\section{Supplemental Materials}
\label{sec:addRelated}

\subsection{Broader Impact}

It is widely accepted that large labeled datasets are an essential component of training deep neural networks, either directly for training or indirectly via transfer learning.
To the best of our knowledge, this paper is the first to demonstrate performance comparable to fully supervised learning with \OSSSLno.
Eliminating the burden of labeling massive amounts of training data creates great potential for new neural network applications that attain high performance, which is especially important when labeling requires expertise.  
Hence, the societal impact will be to make deep learning applications even more widespread.


From a scientific perspective, \OSSSL provides important insights on the intricacies of training deep neural networks.
The effect of changing just one training image can significantly impact the final performance.
Unlike fully supervised learning that commonly deals with the training of large datasets, this method provides a technique to gain information about the impact of a single labeled sample in training.
In addition, we anticipate that further investigation into the instability issues of \OSSSL will lead to new understandings of training neural network.

Furthermore, the experience of training highly sensitive networks provides an educational experience on hyper-parameter tuning that carries over to easier training situations.
In order to achieve convergence with \OSSSLno, one must learn how to tune the hyper-parameters and architecture well.
Similarly, we believe that utilizing \OSSSL with automatic methods such as AutoML and neural architecture search (NAS) will lead to better choices for hyper-parameters and architectures. 

\textbf{Limitations:}
While our work has taken valuable steps towards making one or few-shot \SSL possible for applications, a large gap still remains before this can be realized in practice, especially due to issues with stability during training and hyper-parameter sensitivity.
The sensitivity of the results to choices of the hyper-parameters makes \OSSSL difficult to use in real-world applications.
While there is a wide range of valuable applications (e.g., medical) that could benefit from \SSLno, the testing of these applications is beyond the scope of this work.

While we attempted to provide a thorough investigation, there are a number of limitations in our work and several factors that we did not have sufficient time to explore.
Our implementation was built on \SOTA FixMatch algorithm but the ideas presented here should carry over to other \SSL methods, such as the $\pi$ model, temporal ensembling \cite{laine2016temporal}, VAT \cite{miyato2018virtual}, ICT \cite{verma2019interpolation}, UDA \cite{xie2019unsupervised}, $S^4L$ \cite{zhai2019s4l}, MixMatch \cite{berthelot2019mixmatch}, and Mean Teachers \cite{tarvainen2017mean} but this was not tested because none of these other methods  have demonstrated results with less than 25 labeled examples per class.  
The model used in our experiments was a Wide ResNet-28-2 and the experiments were replicated with ShakeNet \cite{gastaldi2017shake} with analogous results, indicating that our conclusions and insights are independent of model architecture.

In addition, we made use of labeled test data to demonstrate the performance of BOSS.
In practical settings, one has a large unlabeled dataset and one wishes to avoid burdensome manual labeling.
However, the samples in the test dataset are less important than the choices for the class prototypes, so a small test dataset can be quickly created from the ``discards'' when searching for iconic prototypes.
A small test dataset is useful for prototype refinement (i.e., deciding which class prototypes to replace) and it provides the practitioner with useful feedback on the system's performance with a little additional effort.
But even without any test data, one can utilize the pseudo-labeled class counts to decide which class prototypes  should be replaced.


\begin{table*}[tb]
	\begin{center}
		\begin{tabular}{|c|c|c|c|c|c|c|c|c|}
			\hline
			Method  & balance & weight decay & LR & Batch & Momentum & $r_u$ & $\tau$ & $\Delta$   \\
			\hline\hline
			FixMatch & 0 & $ 5 \times 10^{-4} $  & $ 0.03 $  & $ 64 $ & $ 0.88 $ & $ 7 $ & $ 0.95 $ & $ 0 $   \\
			\hline
			Cifar training & 1, 4 & $ 8 \times 10^{-4} $  & $ 0.06 $  & $ 30 $ & $ 0.88 $ & $ 9 $ & $ 0.95 $ & $ 0.25 $  \\
			\hline
			Cifar training & 2, 3 & $ 8 \times 10^{-4} $  & $ 0.06 $  & $ 30 $ & $ 0.88 $ & $ 9 $ & $ 0.9 $ & $ 0 $   \\
			\hline
			Self-training & 4 & $ 5 \times 10^{-4} $  & $ 0.03 $  & $ 64 $ & $ 0.88 $ & $ 7 $ & $ 0.95 $ & $ 0.25 $  \\
			\hline
			SVHN training & 1, 4 & $ 6 \times 10^{-4} $  & $ 0.04 $  & $ 32 $ & $ 0.85 $ & $ 7 $ & $ 0.95 $ & $ 0.25 $  \\
			\hline
			SVHN training & 2, 3 & $ 6 \times 10^{-4} $  & $ 0.04 $  & $ 32 $ & $ 0.85 $ & $ 7 $ & $ 0.9 $ & $ 0 $  \\
			\hline
			Self-training & 0 & $ 6 \times 10^{-4} $  & $ 0.04 $  & $ 32 $ & $ 0.85 $ & $ 7 $ & $ 0.95 $ & $ 0.25 $  \\
			\hline
			
		\end{tabular}
	\end{center}
	\caption{Hyper-parameter values for each of the various steps in the training.
	}
	\label{tab:HP}
\end{table*}

\begin{table*}
	\begin{center}
		\begin{tabular}{|c|c|c|c|c|c|}
			\hline
			&  & \multicolumn{4}{|c|}{BOSS balance method}    \\
			\hline
			WD/LR/BS/$r_u$ & FixMatch & 1 & 2 & 3 & 4     \\
			\hline\hline
			$5 \times 10^{-4}/0.03/64/7$ & $ 74 \pm 5  $ & $ 34 \pm 2  $  &  $ 44 \pm 7  $  & $ 40 \pm 2   $  & $ 31.5  \pm 0.5  $   \\
			\hline
			$8 \times 10^{-4}/0.06/30/9$ & $ 47 \pm 8 $ & $ 93 \pm 0.7 $  &  $ 90 \pm 2 $  & $ 84 \pm 13  $  & $ 78 \pm 20  $   \\
			\hline

		\end{tabular}
	\end{center}
	\caption{Test accuracies for class prototype set 2 for two  hyper-parameter settings. The hyper-parameters are weight decay (WD), learning rate (LR), batch size (BS), and the ratio of the unlabeled to labeled data ($r_u$).
	}
	\label{tab:set2HP}
\end{table*}


Furthermore, there are several assumptions that might not hold true in a practical setting.
First of all, there is an implicit assumption that the unlabeled dataset is class balanced; that is, it contains the same number of samples of each class.
In practical situations with large amounts of unlabeled data, this assumption is unlikely to be true.
In cases where the number of unlabeled samples belonging to each class can be estimated, it is possible to adapt the class balancing methods.
When the number of unlabeled samples belonging to each class is unknown, it is possible to create a small validation set in a similar manner as described above for creating a test set and utilize the validation set as a measure of class balance.

In addition, we also assume in our experiments that all of the unlabeled samples belong to one of the known classes.
In practical settings, the unlabeled dataset might contain samples that don't belong to any of the prototype classes.
We did not test the situation where we use only a subset of the classes in the training datasets.

\subsection{Hyper-parameters}
\label{sec:expHP}

For FixMatch we used the default hyper-parameters that were specified in Sohn, \etal \cite{sohn2020fixmatch}.
However, in our initial experiments with the class balance methods, we found that these hyper-parameters performed poorly.
Therefore, we used a different set of \HP values for FixMatch and for the BOSS methods.


Table \ref{tab:HP} contains the hyper-parameter values used for the results reported in our paper.
Additional hyper-parameter settings that were consistent over all the runs include setting kimgs = 32768 (i.e., the number of training images) and $\lambda_u = 1$ (i.e., the unlabeled loss multiplicative factor).
Furthermore, we set the augment input parameter to `d.d.d', which is the default data augmentation for the labeled and unlabeled data.
Our early experiments with setting the augment input parameter to `d.d.rac' produces small improvements so we subsequently used the default values.
The balance column reflects the class balancing method used (balance = 0 corresponds to FixMatch, which does not use any class balancing method).
The remaining columns specify the weight decay, learning rate, batch size, momentum, ratio of unlabeled to labeled data, confidence threshold, and change in the confidence threshold for minority classes.
Details of these last three hyper-parameters are provided in the main text.

Specifically, we found that increasing the ratio of unlabeled to labeled data (from 7 to 9), weight decay (from $ 5 \times 10^{-4}$ to $8 \times 10^{-4}$)  and the learning rates (from 0.03 to 0.06) improved performance.
We also found that decreasing the confidence threshold from 0.95 to 0.9 improved performance but for class balancing methods 1 and 4, we left the confidence threshold at 0.95 because the class-based thresholds were lowered by these class balancing methods.
We also discovered that a smaller batch-size improved performance and chose a batch size of 30 that was a multiple of the number of classes. 
Our experiments with momentum found a small improvement with values between 0.85 and 0.9 and settled on using 0.88 for our experiments.

As mentioned above, we tried to use the same \HPs for both FixMatch and for the class balancing methods but this proved to provide an unfair comparison to one or the other.
Table \ref{tab:set2HP} illustrates this.
This Table provides the averaged test accuracies for class prototype set 2 for the default and another choice of weight decay (WD), learning rate (LR), batch size (BS), and the ratio of the unlabeled to labeled data ($r_u$).
The results for the BOSS methods improve significantly by tuning the \HPs but the performance of FixMatch is reduced substantially.
So we used the default set of \HPs for FixMatch and another set of \HP values for the class balance methods.

\section{Implementation details}
\label{sec:Implementation}

In this Section we describe the changes we made to the original FixMatch codes and provide guidance on how to replicate our experiments.  This Section relies on the reader being familiar with the TensorFlow version at \url{https://github.com/google-research/fixmatch} and the PyTorch version located at \url{https://github.com/CoinCheung/fixmatch}.  We provide a copy of our codes as part of our Supplemental Materials.

Modifications to the original TensorFlow version of the FixMatch code were localized.
In the TensorFlow version, the primary changes were made to \emph{fixmatch.py}.
This includes the implementation of the four class balancing methods.
In support of these methods, the code for computing the number of pseudo-labels in each class was implemented.
Also, a few new input parameters were added to this file that are related to the class balancing methods.
Specifically, we added the input parameter ``balance'' to specify the class balancing method (balance=0 acts the same as the original FixMatch code) and ``delT'' (i.e., $ \Delta $) as the amount that balance method 1 can reduce the threshold.
Modifications were also made to \emph{cta/lib/train.py} to compute test accuracies for each class, keep track of the best test accuracy, and output the sorted pseudo-labels for the unlabeled training data.
In addition, changes to \emph{libml/data.py} and \emph{libml/augment.py} were required in order to accept the new prototype versions of the labeled datasets.

In addition to the code, the TensorFlow FixMatch version required several other steps that are supported by code in the \emph{scripts} folder.
Instructions for creating the necessary dataset files are located on the website at \url{https://github.com/google-research/fixmatch}.
These instructions use programs in the scripts folder that needed to be modified in order to create the dataset files needed for the prototype sets and for self-training.

We named the prototype datasets with a `p' at the end to distinguish them from the original datasets.
That is, `cifar10' became `cifar10p' and `svhn' became `svhnp'.
Therefore, it was necessary to create \emph{scripts/cifar10\_prototypes.py} and \emph{scripts/svhn\_prototypes.py} to generate the labeled training data files.
We note that to be consistent with the TensorFlow FixMatch, we used `seed' as the input parameter to represent different prototype sets.
It is also necessary to copy the unlabeled training and labeled training files from the cifar10/svhn file names to the cifar10p/svhnp file names and we provide shell scripts to do so.

Self-training is performed as a separate step from the first training run.
The training run will have created three files containing the pseudo-labels for the unlabeled training data sorted from the most confident predictions down. 
The three files are the pseudo-labels, the confidences, and the true labels (used only for debug purposes).
The programs \emph{scripts/cifar10\_iteration.py} and \emph{scripts/svhn\_iteration.py} are provided to combine the highest confidence pseudo-labeled examples with the labeled class prototypes and create the necessary files for the self-training run.
We provide shell scripts as a template for how this is done.
Once these files are created, the self-training iteration can be run.

Most of our experiments were run on a SuperMicro SuperServer with Tesla V100 GPUs.
We discovered that it was important to run our experiments on only 1 GPU and all our runs using multiple GPUs performed poorly.

Modifications to the PyTorch version of the FixMatch code were simpler than for the TensorFlow code.
However, the execution of this code ran almost three times longer, which greatly reduced the number of experiments we could run due to constraints on computational resources.
The primary modifications for class balancing were added to \emph{label\_guessor.py}.
Secondary modification were made to the main program in \emph{train.py} to add the class balancing input parameters and arguments for the call to label\_guessor.
In addition, \emph{cifar.py} was modified to use the class prototypes instead of random examples.
It was not necessary to create class prototype files as it was with the TensorFlow version.
We did not have sufficient time to test self-training with the PyTorch version.


\end{document}